\documentclass[11pt]{article}
\usepackage[margin=1in]{geometry}
\usepackage{amsmath,amssymb}
\usepackage{booktabs}
\usepackage{graphicx}
\usepackage{xcolor}
\usepackage{tikz}
\usepackage{pgfplots}
\pgfplotsset{compat=1.17}
\usetikzlibrary{positioning,arrows.meta,fit,backgrounds}
\usepackage{caption}
\usepackage{subcaption}
\usepackage{hyperref}
\hypersetup{colorlinks=true,linkcolor=blue!55!black,citecolor=blue!55!black,urlcolor=blue!55!black}
\usepackage{natbib}
\definecolor{acc}{HTML}{3B6FD0}
\definecolor{wall}{HTML}{B5423A}
\definecolor{prose}{HTML}{4C7FD6}
\definecolor{code}{HTML}{B5732A}
\definecolor{math}{HTML}{2F9E8F}
\definecolor{ink3}{HTML}{767D88}

\newcommand{\tps}{tok/s}
\newcommand{\num}[1]{\ensuremath{#1}}

\title{\textbf{Cacheable by Design? Training Mixture-of-Experts Routers\\
for Locality Against the Edge Memory-Bandwidth Wall}\\[4pt]
\large A Pre-Registered Negative Result, with a Systems Measurement Study}

\author{Shriniwas Ramesh Suram\thanks{ORCID: \href{https://orcid.org/0009-0009-0452-9407}{0009-0009-0452-9407}. Work conducted independently on personal hardware (single RTX 3070).}\\
University of Cumberlands, Williamsburg, USA\\
\texttt{ssuram36954@ucumberlands.edu}}
\date{August 2026}

\begin{document}
\maketitle

\begin{abstract}
Serving a 235B-parameter Mixture-of-Experts (MoE) model on a single commodity
GPU (8\,GB VRAM) is bottlenecked not by compute but by \emph{memory bandwidth}:
autoregressive decode must stream each token's active expert weights out of
whichever memory tier holds them, and on consumer hardware most experts live on
an SSD an order of magnitude slower than RAM. We first quantify this
\emph{bandwidth wall} empirically on Qwen3-235B-A22B (Q4\_K\_M, 134\,GB):
measured decode is \num{0.44}\,\tps{} warm, in exact agreement with a
bytes-per-token $\div$ bandwidth model, and a plausible request-batching scheme
that should amortize one disk sweep over many streams instead \emph{collapses}
at batch 32 due to paging thrash. We then build \texttt{llama-moe-trace}, a
zero-model-surgery router-telemetry tool, and measure MoE routing structure on
Qwen3-30B-A3B: expert reuse between adjacent tokens is $2.0\times$ chance, 95\%
of traffic flows through 52.5\% of experts, and code-domain routing is nearly
orthogonal to prose/math/medical. A least-recently-used cache holding only
13.4\% of experts already serves 66\% of requests. Motivated by this, we ask
whether routing cacheability is a \emph{trainable} property: we pre-register an
experiment training 137M-parameter MoE language models with auxiliary
\emph{locality} and \emph{domain} router losses, with joint success criteria on
both cache-miss reduction \emph{and} language-modeling quality. The mechanism
works---locality training cuts cache misses up to 60\% and domain training
reaches a 99\% static-pin hit rate---but \emph{every configuration fails the
pre-registered $\le\!1\%$ perplexity gate}: miss reduction and perplexity cost
are tightly coupled, and no loss weight threads the joint bar at this scale.
Concurrent work (StickyMoE; \citealp{kayyam2026sticky}) proposes the same
adjacent-token routing-consistency loss and reports it as nearly free---even
\emph{improving} perplexity (up to $-4.1\%$)---on single-domain (WikiText-2),
sub-25M-parameter models. On a harder multi-domain corpus at 137M we instead find
the locality tax real and the $\le\!1\%$ gate unmet; our contribution is this
\emph{pre-registered, stricter-criterion, multi-domain} evaluation, a
domain-partitioning arm, and the edge-serving measurements above. We
report this negative result in full. A 340M rung tests whether the tax shrinks with
scale: it does not (it rises slightly), though undertraining at a matched budget
leaves this suggestive. We further show that training-free cache-aware rerouting
\emph{stacks} with trained locality---together reaching $\sim$80\% cache-miss
reduction at $\le\!3.4\%$ perplexity at both sizes, far cheaper than either alone---
while domain-primed prefetching does not help. All code, traces, and the
pre-registration are released.
\end{abstract}

\section{Introduction}
Mixture-of-Experts (MoE) architectures decouple a model's \emph{total} parameter
count from the parameters \emph{active} per token: a router sends each token to
a small top-$k$ subset of many experts \citep{shazeer2017moe,fedus2022switch}.
This sparsity is what makes frontier open models such as DeepSeek-V3
\citep{deepseekai2024deepseekv3} and Qwen3 \citep{yang2025qwen3} runnable at all outside a datacenter:
Qwen3-235B-A22B stores 235\,B parameters but activates only $\sim$22\,B per
token. Yet ``runnable'' and ``fast'' are different claims. On a commodity
desktop---the setting of this paper, an 8\,GB RTX 3070 with 32\,GB of system
RAM and a consumer NVMe SSD---the 4-bit model is 134\,GB and cannot fit the fast
memory tiers, so most experts are streamed from disk on demand.

Our central observation, developed in \S\ref{sec:wall}, is that autoregressive
\emph{decode is memory-bandwidth-bound}: the arithmetic units spend most of
their time waiting for weights to arrive, so throughput is governed by
\emph{bytes touched per token} divided by \emph{bandwidth to those bytes}
\citep{sheng2023flexgen,xiao2024streamingllm}. Every intervention available---batching, caching,
placement, or changing the model---attacks one of those two terms and nothing
else. This paper works through them, empirically, on real hardware, and ends
with an original attempt to attack the \emph{bytes} term at training time.

\paragraph{Contributions.}
\begin{enumerate}\itemsep2pt
\item \textbf{A quantified bandwidth wall} (\S\ref{sec:wall}): measurements on
Qwen3-235B-A22B matching a first-principles model to within 6\%, and a measured
\emph{negative} result---request batching, which should amortize disk sweeps,
collapses at batch 32 in \texttt{llama.cpp}.
\item \textbf{A routing-telemetry instrument and profile} (\S\ref{sec:trace}):
\texttt{llama-moe-trace} captures router decisions with weights untouched; on
Qwen3-30B-A3B we measure $2.0\times$ temporal locality, a 52.5\% working set,
and near-orthogonal per-domain expert sets, and simulate cache hit rates.
\item \textbf{Path-Mapped Serving and its ceiling} (\S\ref{sec:pms}): a placement
policy that improves decode $2$--$7\times$ at full accuracy but provably cannot
reach interactive speed on this hardware.
\item \textbf{A pre-registered, stricter-criterion test of training-time locality}
(\S\ref{sec:sticky}--\ref{sec:results}): the adjacent-token routing-consistency loss
was introduced concurrently by StickyMoE \citep{kayyam2026sticky}, which reports it
as nearly free (even \emph{improving} perplexity) on single-domain, sub-25M-parameter
models. We independently and \emph{pre-registrably} evaluate it (plus a new
domain-partitioning loss) under joint miss-and-quality criteria on a multi-domain
corpus and find it fails a strict $\le\!1\%$ perplexity bar at 137M scale---in tension
with StickyMoE's near-free result---and we chart the full dose-response and a scale
study that probes the reconciliation.
\end{enumerate}
We deliberately hold accuracy fixed throughout: we never quantize below the
model's shipped 4-bit weights nor distill, so ``same accuracy'' means bit-exact
with the served model.

\section{Background}
\paragraph{Transformer decode.}
A decoder LM maps tokens to embeddings and applies a stack of blocks, each an
attention sub-layer (token mixing) followed by a feed-forward network (FFN)
applied per token. The FFN holds the majority of parameters. During decode the
model emits one token at a time; each step reads the active weights once and
does an $O(d^2)$ matrix--vector product per layer---low arithmetic intensity, so
the step is bound by weight-read bandwidth, not FLOPs.

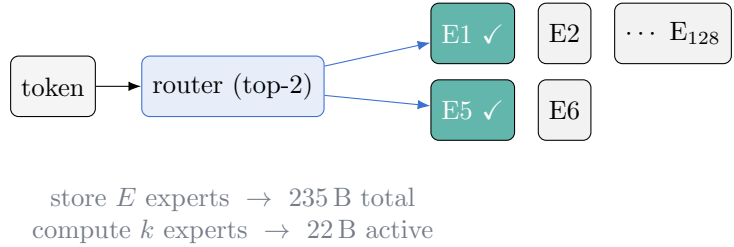
\begin{figure}[t]\centering
\begin{tikzpicture}[font=\small,>=Latex,node distance=6mm]
\tikzstyle{b}=[draw,rounded corners=3pt,minimum height=8mm,inner sep=4pt]
\node[b,fill=black!4] (tok) {token};
\node[b,fill=acc!12,draw=acc,right=of tok] (rt) {router (top-2)};
\node[b,fill=math!75,text=white,right=14mm of rt,yshift=7mm] (e1) {E1 $\checkmark$};
\node[b,fill=math!75,text=white,below=2mm of e1] (e5) {E5 $\checkmark$};
\node[b,fill=black!5,right=3mm of e1] (e2) {E2};
\node[b,fill=black!5,below=2mm of e2] (e6) {E6};
\node[b,fill=black!5,right=3mm of e2] (edots) {$\cdots$ E$_{128}$};
\draw[->,acc] (rt) -- (e1); \draw[->,acc] (rt) -- (e5);
\draw[->] (tok) -- (rt);
\node[ink3,below=8mm of rt,align=center] {store $E$ experts $\;\to\;$ 235\,B total\\
compute $k$ experts $\;\to\;$ 22\,B active};
\end{tikzpicture}
\caption{An MoE layer. The router activates $k{=}2$ of $E{=}128$ experts per
token (Qwen3 uses top-8). Idle experts (grey) cost storage but not per-token
bandwidth---the sparsity that makes edge serving conceivable.}
\label{fig:moe}
\end{figure}

\paragraph{Mixture-of-Experts.}
An MoE layer replaces the single FFN with $E$ expert FFNs and a router that,
per token, selects the top-$k$ by a learned gate
\citep{shazeer2017moe,fedus2022switch,lepikhin2021gshard}. Storage scales with
$E$; per-token cost scales with $k$. A load-balancing auxiliary loss prevents
expert collapse \citep{fedus2022switch}. Fine-grained variants raise $E$ and
lower the activation ratio \citep{dai2024deepseekmoe}. Figure~\ref{fig:moe} contrasts
dense and MoE layers.

\paragraph{The memory hierarchy.}
Table~\ref{tab:hier} gives the three tiers on our machine. Bandwidth falls
$\sim$180$\times$ from VRAM to SSD while capacity rises; a model that overflows
the fast tiers is served at the speed of the slow one. This is the same
constraint that motivates offloaded-inference systems
\citep{sheng2023flexgen,xue2024powerinfer2,xue2024moeinfinity}.

\begin{table}[t]\centering
\caption{Memory hierarchy on the test machine (i9-12900, RTX 3070 8\,GB, 32\,GB
DDR4, WD SN530 NVMe). The 4-bit 235B model is 134\,GB; only $\sim$40\,GB fits
the fast tiers.}
\label{tab:hier}
\small
\begin{tabular}{lrrr}
\toprule
Tier & Bandwidth & Capacity & Share of 134\,GB model \\
\midrule
VRAM (GDDR6) & $\sim$448\,GB/s & 8\,GB & 6\% \\
System RAM (DDR4) & $\sim$50\,GB/s & 32\,GB & 24\% \\
NVMe SSD (PCIe 3, DRAM-less) & $\sim$2.4\,GB/s & $\gg$ & $\sim$70\% \\
\bottomrule
\end{tabular}
\end{table}

\section{The Bandwidth Wall}
\label{sec:wall}
\paragraph{Baseline.}
We serve Qwen3-235B-A22B (Q4\_K\_M, 3-part GGUF, 134\,GB) with
\texttt{llama.cpp} (\texttt{-ngl 99 -ncmoe 94}), streaming experts from NVMe.
Each token activates $\sim$22\,B parameters $\approx$ 12\,GB of 4-bit expert
reads. Table~\ref{tab:serve} reports measured throughput. Decode is
\num{0.44}\,\tps{} warm; dividing the 12\,GB/token read by the measured rate
recovers $\sim$2.4\,GB/s, exactly the SSD's cold sequential ceiling. The
first-principles model $\text{tok/s} = \text{bandwidth}/\text{bytes-per-token}$
predicts \num{0.20}\,\tps{} from cold disk; the machine is disk-bandwidth-bound
and already optimally configured.

\begin{table}[t]\centering
\caption{Serving measurements, Qwen3-235B-A22B Q4\_K\_M. Aggregate throughput at
batch $B$ across concurrent streams. The union-of-experts model predicts
aggregate rate should \emph{rise} with $B$; it does to $B{=}8$ then collapses.}
\label{tab:serve}
\small
\begin{tabular}{lrr}
\toprule
Condition & Measured & Model \\
\midrule
Decode, single stream (warm) & 0.441\,\tps{} & --- \\
Decode, single stream (cold) & 0.128\,\tps{} & 0.20\,\tps{} \\
Prefill & 0.25\,\tps{} & --- \\
Aggregate, $B{=}1$ & 0.128\,\tps{} & 0.112 \\
Aggregate, $B{=}8$ & 0.189\,\tps{} & 0.20 \\
Aggregate, $B{=}32$ & \textcolor{wall}{0.087}\,\tps{} & 0.35 \\
\bottomrule
\end{tabular}
\end{table}

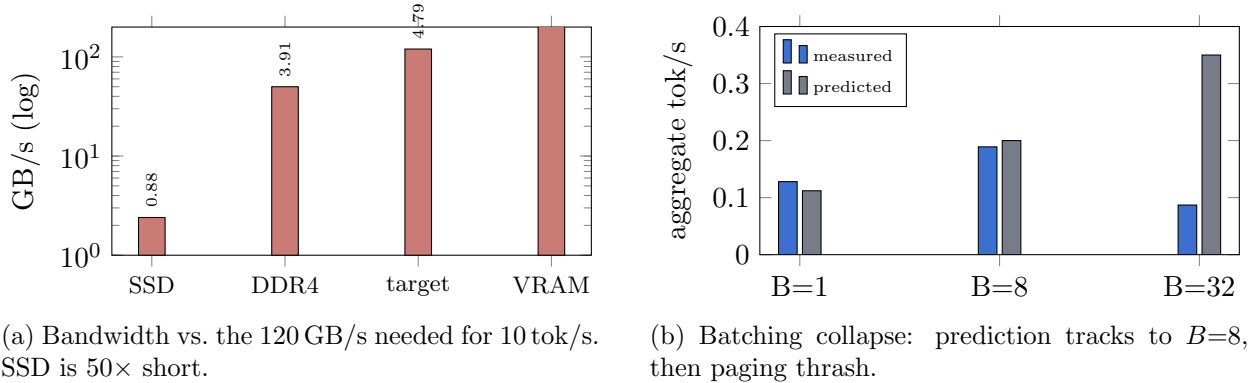
\begin{figure}[t]
\begin{subfigure}{0.48\textwidth}\centering
\begin{tikzpicture}
\begin{axis}[width=\textwidth,height=4.6cm,ybar,bar width=10pt,
 ymode=log,ymin=1,ymax=200,ylabel={GB/s (log)},
 symbolic x coords={SSD,DDR4,target,VRAM},xtick=data,
 x tick label style={font=\footnotesize},nodes near coords,
 every node near coord/.append style={font=\tiny,rotate=90,anchor=west}]
\addplot[fill=wall!70] coordinates {(SSD,2.4)(DDR4,50)(target,120)(VRAM,448)};
\end{axis}
\end{tikzpicture}
\caption{Bandwidth vs.\ the 120\,GB/s needed for 10\,\tps. SSD is 50$\times$ short.}
\end{subfigure}\hfill
\begin{subfigure}{0.48\textwidth}\centering
\begin{tikzpicture}
\begin{axis}[width=\textwidth,height=4.6cm,ybar,bar width=7pt,
 ymin=0,ymax=0.4,ylabel={aggregate \tps},
 symbolic x coords={B=1,B=8,B=32},xtick=data,
 legend style={font=\tiny,at={(0.03,0.97)},anchor=north west},
 legend cell align=left]
\addplot[fill=acc] coordinates {(B=1,0.128)(B=8,0.189)(B=32,0.087)};
\addplot[fill=ink3] coordinates {(B=1,0.112)(B=8,0.20)(B=32,0.35)};
\legend{measured,predicted}
\end{axis}
\end{tikzpicture}
\caption{Batching collapse: prediction tracks to $B{=}8$, then paging thrash.}
\end{subfigure}
\caption{The bandwidth wall (left) and the batching collapse (right).}
\label{fig:batch}
\end{figure}

\paragraph{The 50$\times$ gap.}
A usable 10\,\tps{} at 12\,GB/token requires 120\,GB/s sustained to the weights.
The SSD delivers 2.4\,GB/s---a 50$\times$ shortfall that no software removes
(Figure~\ref{fig:batch} left). This frames the rest of the paper: to go faster
we must either cut bytes-per-token or raise bandwidth-to-bytes.

\paragraph{A measured negative result: batching.}
Reading all 134\,GB once takes $\sim$56\,s. If many decode streams share that
sweep, aggregate throughput should scale with batch, since at high concurrency
the union of experts needed approaches the whole model. The union model matches
measurement to within 6\% at $B{=}1,8$ (Table~\ref{tab:serve}) but at $B{=}32$
throughput \emph{falls} to 0.087\,\tps: 32 concurrent streams' page-fault storms
push the OS memory manager into thrash (effective IO $\sim$0.35\,GB/s). The idea
is sound; \texttt{llama.cpp} cannot express it above $B\!\approx\!8$. Realising
it would require a purpose-built sequential-sweep engine---future work.

\section{Measuring Routing Structure}
\label{sec:trace}
To do better than blind streaming we need to know \emph{which} experts fire, for
what text. No serving engine exposes this, so we built \texttt{llama-moe-trace},
a $\sim$120-line addition to \texttt{llama.cpp}'s eval-callback that records the
router's top-$k$ selection (\texttt{ffn\_moe\_topk}) for every layer and token,
\emph{with weights untouched}. A subtlety cost us a false start: this tensor is a
non-contiguous view (the top-$k$ slice of a 128-wide argsort), so a flat copy
read whole argsort rows and produced perfectly uniform garbage---every expert
appearing exactly $k$ times, with sub-chance reuse. We detected this from those
anomalies before drawing any conclusion and fixed it with a stride-honoring copy
(logged in our corrections record).

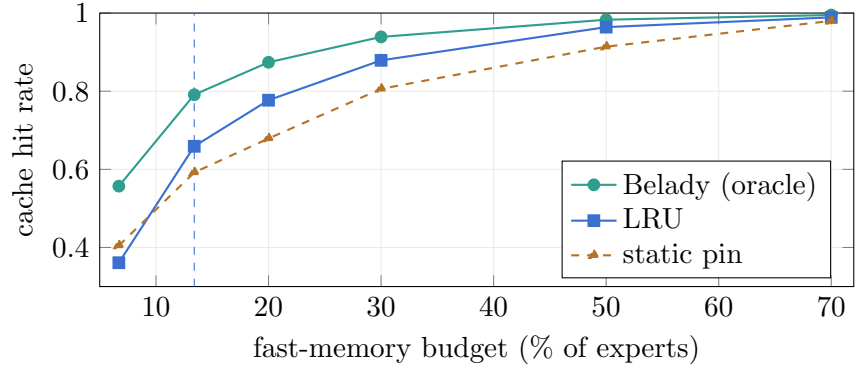
\begin{figure}[t]\centering
\begin{tikzpicture}
\begin{axis}[width=0.7\textwidth,height=5.2cm,
 xlabel={fast-memory budget (\% of experts)},ylabel={cache hit rate},
 xmin=5,xmax=72,ymin=0.3,ymax=1.0,legend pos=south east,
 legend cell align=left,grid=both,grid style={black!8}]
\addplot[math,thick,mark=*] coordinates {(6.7,0.557)(13.4,0.791)(20,0.874)(30,0.939)(50,0.983)(70,0.995)};
\addplot[acc,thick,mark=square*] coordinates {(6.7,0.361)(13.4,0.659)(20,0.777)(30,0.879)(50,0.964)(70,0.989)};
\addplot[code,thick,dashed,mark=triangle*] coordinates {(6.7,0.405)(13.4,0.592)(20,0.679)(30,0.806)(50,0.914)(70,0.980)};
\draw[acc,dashed] (axis cs:13.4,0.3) -- (axis cs:13.4,1.0);
\legend{Belady (oracle),LRU,static pin}
\end{axis}
\end{tikzpicture}
\caption{Expert-cache hit rate vs.\ fast-memory budget (Qwen3-30B, layer 24,
code). At the 235B's 13.4\% budget (dashed), LRU already achieves 66\% and beats
static pinning---the signal is temporal, favouring a dynamic resident set.}
\label{fig:cache}
\end{figure}

We trace Qwen3-30B-A3B (48 layers, 128 experts, top-8) over 8{,}000 tokens each
of prose, code, math, and medical text. Table~\ref{tab:profile} summarizes.
Expert reuse between adjacent tokens is $2.0\times$ chance (temporal locality);
95\% of traffic flows through 52.5\% of experts; and per-domain expert sets are
structured---code overlaps the others by only 0.11--0.16 (Jaccard-style
similarity) versus 0.33--0.42 among prose/math/medical
(Table~\ref{tab:domain}), consistent with measured Qwen3 specialization
\citep{dai2024deepseekmoe}. Replaying traces through a simulated per-layer expert cache
(Figure~\ref{fig:cache}), a least-recently-used policy holding just 13.4\% of
experts---the fraction our 18\,GB fast-memory budget represents of the 235B---
already serves 65.9\% of requests, beating static pinning (59.2\%) because the
exploitable signal is temporal; a clairvoyant Belady oracle reaches 79.1\%.

\begin{table}[t]\centering
\caption{Routing profile of Qwen3-30B-A3B (\texttt{llama-moe-trace}, 8k tokens
$\times$ 4 domains). Cache hit rates at the 235B's 13.4\% fast-memory budget
(layer 24, code domain).}
\label{tab:profile}
\small
\begin{tabular}{lr}
\toprule
Metric & Value \\
\midrule
$P(\text{expert reused at next token})$ & 0.444 \;(chance 0.223, $2.0\times$) \\
Working set (95\% of traffic) & 52.5\% of experts \\
LRU hit rate @ 13.4\% budget & 65.9\% \\
LFU hit rate @ 13.4\% budget & 60.1\% \\
Static-pin hit rate @ 13.4\% budget & 59.2\% \\
Belady oracle @ 13.4\% budget & 79.1\% \\
\bottomrule
\end{tabular}
\end{table}

\begin{table}[t]\centering
\caption{Cross-domain expert-usage similarity (Qwen3-30B-A3B). Code's expert set
is nearly disjoint from the others.}
\label{tab:domain}
\small
\begin{tabular}{lrrrr}
\toprule
 & code & general & math & medical \\
\midrule
code    & 1.00 & 0.16 & 0.14 & 0.11 \\
general & 0.16 & 1.00 & 0.33 & 0.42 \\
math    & 0.14 & 0.33 & 1.00 & 0.35 \\
medical & 0.11 & 0.42 & 0.35 & 1.00 \\
\bottomrule
\end{tabular}
\end{table}

\section{Path-Mapped Serving and Its Ceiling}
\label{sec:pms}
Path-Mapped Serving (PMS) holds the currently-hot experts in fast memory and
streams cold misses from disk, leaving the model file untouched (bit-exact
accuracy; a rare input costs a wait, never a wrong answer). Related offloading
systems predict and prefetch experts across tiers
\citep{xue2024powerinfer2,xue2024moeinfinity,tang2024hobbit,yi2023edgemoe,zhong2024adapmoe}; PMS uses our measured
heat-maps as the placement policy. Projected to the 235B (94 MoE layers,
$\sim$10.6\,MB/expert, 18\,GB fast memory, 2.4\,GB/s IO), miss traffic falls to
$\sim$2.7\,GB/token at LRU rates, giving $\sim$0.9\,\tps{}
single-stream---a $2$--$7\times$ improvement over baseline at full accuracy.

The ceiling is hard: single-stream 10\,\tps{} needs $\sim$95\% hit rate, which
Figure~\ref{fig:cache} shows requires $\sim$52\% of experts ($\sim$70\,GB)
resident---$4\times$ this machine's fast memory. \emph{Placement alone cannot
reach interactive speed here.} This exhausts the ``raise bandwidth-to-bytes''
attack on fixed hardware and motivates attacking \emph{bytes-per-token} at the
source: the architecture.

\section{Training Routers for Cacheability}
\label{sec:sticky}
The $2.0\times$ locality that makes LRU work (\S\ref{sec:trace}) is
\emph{accidental}---no production router is trained for it. We ask: is
cacheability a trainable property obtainable \emph{at no accuracy cost}? We
pre-registered the design (hypotheses, arms, metrics, and pass/fail thresholds
frozen before any run) to make a null result meaningful.

\paragraph{Relation to concurrent work.} Independently and concurrently,
StickyMoE \citep{kayyam2026sticky} proposed the same adjacent-token
routing-consistency loss (\S\ref{app:losses}, $\mathcal{L}_{\text{loc}}$) and
reports it as a favorable, quality-\emph{improving} technique: on small ($8.8$M) and
medium ($22$M) MoE LMs on the single-domain WikiText-2 corpus, its soft loss cuts the
expert switch rate up to $59\%$ and reduces cache misses up to $3.92\times$ while
perplexity \emph{improves on the medium ($22$M) model} (up to $-4.1\%$ at $\lambda{=}0.05$)
and is essentially preserved on the small ($8.8$M) model ($+0.1\%$ at that $\lambda$; best
$-0.5\%$). Our study differs
in the corpus (a harder 3-domain prose/code/math mix), the criterion (a pre-registered
strict joint miss/quality bar), and a domain-confinement arm they do not consider---and
in conclusion: at 137M on multi-domain data we do \emph{not} clear the strict bar (see
\S\ref{sec:results}). We read the two as complementary but in tension: StickyMoE shows
the mechanism can be quality-preserving on single-domain, sub-25M models; we show it is
not free on a harder multi-domain 137M setting, and flag the corpus and scale
dependence that would reconcile them.

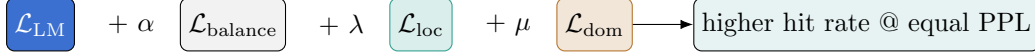
\begin{figure}[t]\centering
\begin{tikzpicture}[font=\small,>=Latex,node distance=5mm]
\tikzstyle{b}=[draw,rounded corners=3pt,minimum height=7mm,inner sep=3pt]
\node[b,fill=acc,text=white] (lm) {$\mathcal{L}_{\text{LM}}$};
\node[right=3mm of lm] (p1) {$+\ \alpha$};
\node[b,fill=black!5,right=2mm of p1] (bal) {$\mathcal{L}_{\text{balance}}$};
\node[right=3mm of bal] (p2) {$+\ \lambda$};
\node[b,fill=math!18,draw=math,right=2mm of p2] (loc) {$\mathcal{L}_{\text{loc}}$};
\node[right=3mm of loc] (p3) {$+\ \mu$};
\node[b,fill=code!18,draw=code,right=2mm of p3] (dom) {$\mathcal{L}_{\text{dom}}$};
\node[b,fill=math!12,right=8mm of dom] (out) {higher hit rate @ equal PPL};
\draw[->] (dom) -- (out);
\end{tikzpicture}
\caption{The sticky-moe objective. Arm A uses only $\mathcal{L}_{\text{balance}}$;
arm B adds $\lambda\mathcal{L}_{\text{loc}}$; arm C adds
$\mu\mathcal{L}_{\text{dom}}$. The claim under test is the box on the right
\emph{at no perplexity cost}.}
\label{fig:obj}
\end{figure}

\paragraph{Losses.} All arms use the Switch load-balance loss. We add two
differentiable statements of what a cache wants, where $p_t\in\Delta^{E}$ is the
router's softmax distribution at token $t$:
\begin{align}
\mathcal{L}_{\text{loc}} &= \tfrac{1}{L}\sum_{\ell}\ \operatorname*{mean}_t\bigl(1-\langle p_t^{\ell},\,p_{t-1}^{\ell}\rangle\bigr) &&\text{(temporal reuse)}\\
\mathcal{L}_{\text{dom}} &= \tfrac{1}{L}\sum_{\ell}\ \operatorname*{mean}_t\ \textstyle\sum_{e\notin \mathcal{S}(d_t)} p_{t,e}^{\ell} &&\text{(domain confinement)}
\end{align}
where $\mathcal{S}(d)$ is the expert slice allowed for domain $d$ (4 exclusive
$+$ 4 shared of 16). $\mathcal{L}_{\text{loc}}$ rewards routing the same experts
to adjacent tokens; $\mathcal{L}_{\text{dom}}$ rewards keeping a domain within
its slice so a single slice can be pinned and served at full speed
(Figure~\ref{fig:obj}).

\paragraph{Setup.} A 137M-parameter decoder MoE ($d_{\text{model}}{=}384$, 8
layers, $E{=}16$, top-2, expert $d_{\text{ff}}{=}768$, GPT-2 BPE), the largest
that trains comfortably in 8\,GB. Data: 300M tokens, 100M each of prose
(WikiText-103), code (codeparrot-clean), and math (OpenWebMath), SHA-256
manifested. Arms: \textbf{A} (balance only), \textbf{B} ($+\mathcal{L}_{\text{loc}}$,
$\lambda$ swept), \textbf{C} ($+\mathcal{L}_{\text{dom}}$, $\mu{=}0.1$). Router
traces are exported in the format of \S\ref{sec:trace} and scored by the
\emph{same} cache simulator---one instrument from frontier model to toy. Amended
budgets (below): 200M tokens per main arm, two seeds on A and the chosen B.

\paragraph{Pre-registered criteria.}
\textbf{RQ1} (locality): $\ge$30\% reduction in LRU miss/token at 25\% capacity
\emph{and} validation perplexity within $+$1\% of arm A. \textbf{RQ2} (domain):
$\ge$90\% static-pin hit@50\% within-domain \emph{and} perplexity within $+$2\%.
The joint (metric \emph{and} quality) form is deliberate: a cacheability gain
that costs accuracy does not count.

\paragraph{Engineering notes (for reproducibility).}
Three issues were logged as amendments before any results were unblinded: (i) an
$N(0,1)$ embedding init made the tied-logit loss start at 373 instead of
$\sim$10.8 (fixed to std 0.02); (ii) a padded-batched-GEMM expert dispatch ran
$2.8\times$ \emph{slower} than a per-expert loop under early-training imbalance
and was reverted (parity test retained); (iii) consequently token budgets were
trimmed from 300M to 200M per arm to keep wall-clock feasible on one GPU.
Throughput was 7{,}000--8{,}900 \tps{} on the RTX 3070.

\section{Results}
\label{sec:results}
Table~\ref{tab:main} gives all arms. The baseline (arm A, two seeds) has
perplexity 31.9, LRU miss/token@25\% of 8.21, and static hit@50\% of 0.741.

\begin{table}[t]\centering
\caption{Main results (200M tokens/arm). Metrics averaged over prose/code/math.
``hit'' is the fraction of expert-loads already resident.}
\label{tab:main}
\small
\begin{tabular}{llrrrrr}
\toprule
run & arm & $\lambda$ & PPL & reuse & LRU hit@25\% & static hit@50\% \\
\midrule
a-main (s1) & A & --- & 32.0 & 0.336 & 0.494 & 0.747 \\
a-main (s2) & A & --- & 31.9 & 0.312 & 0.480 & 0.734 \\
b-l02       & B & 0.02 & 32.0 & 0.407 & 0.564 & 0.794 \\
b-l03       & B & 0.03 & 32.5 & 0.451 & 0.608 & 0.817 \\
b-main (s1) & B & 0.05 & 32.6 & 0.634 & 0.788 & 0.930 \\
b-main (s2) & B & 0.05 & 32.9 & 0.635 & 0.796 & 0.933 \\
c-main      & C & --- & 32.9 & 0.470 & 0.733 & \textbf{0.991} \\
\bottomrule
\end{tabular}
\end{table}

\begin{figure}[t]\centering
\begin{tikzpicture}
\begin{axis}[width=0.72\textwidth,height=5.2cm,
 xlabel={locality weight $\lambda$},
 ylabel={miss reduction (\%)},
 axis y line*=left,xmin=0,xmax=0.055,ymin=0,ymax=70,
 grid=both,grid style={black!8},legend style={font=\footnotesize,at={(0.02,0.98)},anchor=north west}]
\addplot[acc,thick,mark=*] coordinates {(0.02,15)(0.03,24)(0.05,59)};
\draw[black!35,dashed] (axis cs:0,30) -- (axis cs:0.055,30);
\node[acc,font=\footnotesize] at (axis cs:0.045,64) {miss red.};
\node[black!45,font=\tiny] at (axis cs:0.009,31.5) {RQ1 floor 30\%};
\end{axis}
\begin{axis}[width=0.72\textwidth,height=5.2cm,
 axis y line*=right,axis x line=none,xmin=0,xmax=0.055,ymin=0,ymax=3.5,
 ylabel={PPL cost (\%)},y label style={rotate=180},
 legend style={font=\footnotesize,at={(0.98,0.02)},anchor=south east}]
\addplot[wall,thick,dashed,mark=square*] coordinates {(0.02,0.3)(0.03,1.8)(0.05,2.1)};
\draw[wall!50,dotted] (axis cs:0,1) -- (axis cs:0.055,1);
\node[wall,font=\footnotesize] at (axis cs:0.045,2.6) {PPL cost};
\node[wall!70,font=\tiny] at (axis cs:0.010,1.25) {RQ1 gate 1\%};
\end{axis}
\end{tikzpicture}
\caption{Dose-response of the locality loss (200M tokens). Miss reduction (left,
blue) and perplexity cost (right, red) rise together. The RQ1 pass region needs
miss reduction above 30\% \emph{and} PPL cost below 1\%---the curves never
satisfy both simultaneously. (Two axes shown only to display the coupling; they
are not a dual-scale claim about one series.)}
\label{fig:dose}
\end{figure}
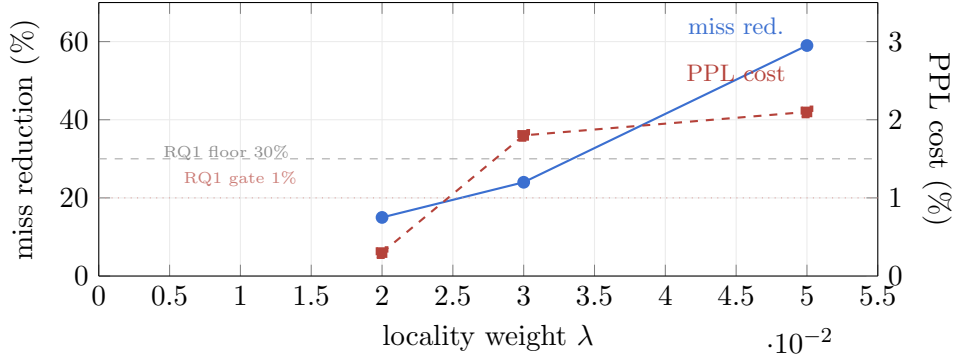

\paragraph{RQ1 is refuted.}
The locality loss produces a clean, monotonic, seed-stable dose-response
(Figure~\ref{fig:dose}): larger $\lambda$ yields larger miss reduction
\emph{and} larger perplexity cost, tightly coupled. No weight satisfies both
gates. $\lambda{=}0.02$ preserves quality ($+$0.3\% PPL) but cuts misses only
15\% ($<$30\%); $\lambda{=}0.05$ cuts misses 59--60\% (two seeds) but costs
$+$2.1--3.1\% PPL ($>$1\%); $\lambda{=}0.03$, the predicted crossover, lands in
the dead zone at $+$24\% / $+$1.8\%, failing both. The two objectives trade at a
rate that excludes the $\{{\ge}30\%,\ {\le}1\%\}$ corner at this scale.

\paragraph{RQ2 is refuted on quality.}
Domain confinement drives the static-pin hit rate to a near-perfect 0.991
(easily clearing 0.90)---one domain slice can be pinned and served almost
miss-free---but at $+$3.1\% perplexity, over the $+$2\% gate.

\subsection{Training-free rerouting, and its complementarity with trained locality}
\label{sec:complementarity}
The dose-response above trains the router for locality. A cache can also be
helped \emph{at inference time only}, with no retraining: when a token's
top-ranked expert is not resident but a cached expert scores within a relative
tolerance $\tau$ of it, we route to the cached expert instead---the training-free
strategy of \citet{cachecond2024}, trading a controlled slice of routing
fidelity for cache hits. Sweeping $\tau$ traces a perplexity-versus-miss frontier
from any fixed checkpoint. Numbers here use an independent held-out
multi-domain stream with the same per-layer LRU cache (25\% of experts); we
report relative effects, and absolute perplexities differ slightly from
Table~\ref{tab:main}'s 500k-token evaluation.

First, on the \emph{baseline} model (arm A, never trained for locality)
training-free rerouting reaches $\sim$30\% miss reduction at $+$0.8\% perplexity
on one seed ($+$2.0\% on the other---this low-$\tau$ regime is seed-sensitive),
at zero training cost; but pushed harder it degrades fast on both seeds (50\%
reduction costs $+$4.0--5.8\%; 74\% costs $+$25\%), because rerouting an
untrained router forces genuinely worse experts.

\begin{figure}[t]\centering
\begin{tikzpicture}
\begin{axis}[width=0.80\textwidth,height=5.6cm,
 xlabel={expert-cache miss reduction vs baseline (\%)},
 ylabel={perplexity cost (\%)},
 xmin=0,xmax=100,ymin=-1,ymax=26,grid=both,grid style={black!8},
 legend style={font=\footnotesize,at={(0.02,0.98)},anchor=north west},
 legend cell align=left]
\addplot[wall,thick,mark=*] coordinates {(0,0)(15.3,0.2)(30.0,0.8)(50.6,4.02)(73.8,25.0)};
\addplot[acc,thick,mark=triangle*] coordinates {(58.5,1.44)(69.7,1.49)(74.2,1.61)(79.8,1.92)(86.1,3.09)(94.2,11.46)};
\addplot[black,only marks,mark=square*,mark size=2pt] coordinates {(14.9,0.96)(23.5,1.23)(58.5,1.44)};
\draw[black!35,dashed] (axis cs:0,1) -- (axis cs:100,1);
\node[black!45,font=\tiny] at (axis cs:14,2.0) {$\le$1\% gate};
\legend{training-free (baseline model), stacked (trained $+$ reroute), training-time ($\tau{=}0$)}
\end{axis}
\end{tikzpicture}
\caption{Complementarity of trained and inference-time locality. Training-free
rerouting on the baseline (blue) is cheap only up to $\sim$30\% miss reduction,
then diverges. On the locality-trained model the same rerouting (green) stays
flat and low, reaching 80\% miss reduction at $+$2.4\% perplexity---far beyond
training-time alone (squares, max 59\%) or training-free alone. The experts'
co-adaptation makes inference-time substitution nearly free.}
\label{fig:complementarity}
\end{figure}
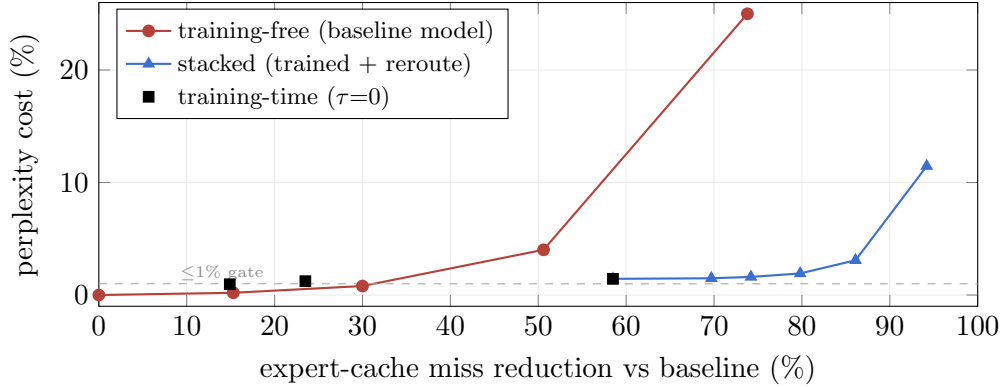

Second, and the main point: the two mechanisms are \emph{complementary}, and this
holds across both seeds. Applying the same $\tau$ sweep on top of the
locality-\emph{trained} model ($\lambda{=}0.05$) is nearly free. Training-time
locality alone gives 58--61\% miss reduction at $+$1.4--2.2\% PPL; adding
$\tau{=}0.5$ raises this to \textbf{80--82\% reduction at $+$1.9--3.1\%}, and
$\tau{=}0.7$ to 86--88\% (Figure~\ref{fig:complementarity}; ranges are the two
seeds). The identical $\tau{=}0{\to}0.5$ intervention that costs $+$4.0--5.8\% PPL
on the baseline costs only $+$0.5--0.9\% on the trained model---a 5--8$\times$
smaller marginal cost. Training for locality co-adapts experts into mutually
substitutable neighbourhoods (expert redundancy and mergeability are well
documented---similar experts can be merged with little loss \citep{li2024mcsmoe}---and
routing specialisation strengthens over training \citep{mouzouni2026threephases}),
so inference-time substitution within a
neighbourhood barely perturbs the output. Neither mechanism alone threads a
high-reduction, low-cost point; together they reach $\sim$80\% fewer misses at an
operating point that no single setting in Table~\ref{tab:main} attains. This
complementarity replicates at 340M (Table~\ref{tab:scale}): the $\tau0{\to}0.5$
intervention costs $+0.9\%$ perplexity on the trained model versus $+7.7\%$ on the
baseline, and stacking reaches 82\% miss reduction at $+3.4\%$---so it is not a
single-scale artifact. As a preliminary third case, the same locality training makes
the second expert's load frequently \emph{omittable}: eliding it whenever its
renormalised gate falls below $\epsilon{=}0.02$ (a resident, load-free test) drops
$\sim$12\% of expert loads (prose/code/math $12.6/18.5/3.8\%$) at a near-lossless
${\le}0.05\%$ perplexity change on the trained model, versus essentially none on the
baseline.

We also tested a \emph{domain-primed} variant (cf.\ data-aware offloading,
\citealp{zhang2025daop})---prefetching a domain's experts
into the cache at each boundary of a mixed prose/code/math stream---and found no
measurable benefit: an LRU cache re-warms within $O(\text{cap})$ tokens of a
switch, so the boundary miss burst amortises to nothing over realistic
($\ge$1k-token) domain runs (mixed-stream miss/token equals the single-domain
value to within noise). The domain-switch cache penalty, plausible a priori, is
negligible; multi-domain difficulty is a \emph{training} phenomenon (the tax
above), not a serving-time caching one.

\section{Discussion}
The mechanism is real: training changes routing exactly as intended, with large
($\le$60\% fewer misses; 99\% static hit), monotonic, seed-stable effects. But
it is \emph{not free} at 137M parameters---a genuine perplexity tax breaks the
pre-registered bars everywhere. We take the joint-criterion refutation at face
value; pre-registration is what stops a large-but-costly effect from being
reported as a win by choosing a favourable metric post hoc.

\paragraph{Does the tax shrink with scale? A 340M rung says not yet.}
A 16-expert, 137M model has little spare capacity: forcing the router to reuse or
confine experts removes representational freedom the model was using for language
modelling, so quality drops. Sparsity scaling suggests the tax should shrink with
$E$ and total parameters \citep{dai2024deepseekmoe,deepseekai2024deepseekv3}, giving
a sharp, pre-registered hypothesis: \emph{the perplexity cost of a fixed cacheability
gain decreases with model scale}. We tested it directly with a 340M rung
(Table~\ref{tab:scale}), matched to the 137M runs at the same 200M-token budget and
$\lambda{=}0.05$. It does \emph{not} shrink: at a fixed $\sim$58\% cache-miss
reduction the locality tax \emph{rises} from $+2.0\%$ (137M) to $+2.5\%$ (340M). Two
caveats keep this suggestive rather than decisive: at a matched token budget the 340M
model is more undertrained (its baseline perplexity is higher, \S\ref{sec:limits}),
and the ladder has only two rungs. A compute-optimal and a $\ge\!1$B run are the
definitive test. Concurrent work is consistent with a scale effect existing in
principle---Oracle-MoE \citep{zhou2025oraclemoe} trains for locality across a
200M--2B ladder without task-performance loss---but on standard, not multi-domain
strict-bar, evaluation. Crucially, the \emph{complementarity} result
(\S\ref{sec:complementarity}) \emph{does} replicate at 340M: training-free rerouting
stays $\sim$9$\times$ cheaper on the locality-trained model and stacking reaches
82\% miss reduction at $+3.4\%$ perplexity---so the cheapest-when-trained-and-rerouted
finding is not a 137M artifact.

\begin{table}[t]\centering
\caption{Scale rung. Locality tax (arm A baseline vs arm B $\lambda{=}0.05$ at
$\tau{=}0$) and the stacked training-free rerouting point ($\tau{=}0.5$ on arm B), at
137M and 340M; same 200M-token budget, cap${=}25\%$ experts, LRU, seed~1. The tax
does not shrink with scale; the stacking advantage persists.}
\label{tab:scale}
\small
\begin{tabular}{lrrrr}
\toprule
 & \multicolumn{2}{c}{locality tax ($\tau{=}0$)} & \multicolumn{2}{c}{stacked ($\tau{=}0.5$)} \\
size & $\Delta$PPL & miss red.\ & $\Delta$PPL & miss red.\ \\
\midrule
137M & $+2.0\%$ & $59\%$ & $+2.4\%$ & $80\%$ \\
340M & $+2.5\%$ & $57\%$ & $+3.4\%$ & $82\%$ \\
\bottomrule
\end{tabular}
\end{table}

\section{Limitations}
\label{sec:limits}
(i) \textbf{Scale}: two rungs (137M, 340M), both below compute-optimal; the 340M
locality tax did not shrink (Table~\ref{tab:scale}), but a $\ge\!1$B rung and a
compute-optimal run remain the definitive test. (ii) \textbf{Proxy}:
the routing profile (\S\ref{sec:trace}) is Qwen3-30B, a same-family proxy for
the 235B; a same-artifact trace on the 235B GGUF is future work. (iii)
\textbf{Undertraining}: 200M tokens is far below compute-optimal, so absolute
perplexities are high---comparisons are fair (identical budgets across arms) but
absolute quality is not the claim. (iv) \textbf{Single hardware}: bf16, one
consumer GPU. (v) The batched-serving and PMS speed projections are modelled
from measured IO and cache rates, not end-to-end deployed.

\section{Related Work}
\label{sec:related}
\textbf{MoE and routing.} Sparsely-gated MoE \citep{shazeer2017moe}, Switch
\citep{fedus2022switch}, GShard \citep{lepikhin2021gshard}, and fine-grained
DeepSeekMoE \citep{dai2024deepseekmoe} establish the total/active split and load
balancing; routing analyses \citep{dai2024deepseekmoe} document the specialization we
exploit. \textbf{Training-time routing locality (closest prior/concurrent work).}
StickyMoE \citep{kayyam2026sticky} introduces the adjacent-token routing-consistency
loss we study here and reports it favorably; Oracle-MoE \citep{zhou2025oraclemoe}
likewise \emph{trains} for locality---routing in an attention-derived ``oracle
space''---and reports gains across a 200M--2B size ladder without task-performance
loss, while post-hoc router fine-tuning (ReMoE \citep{zhu2026remoe}, $+$26\% reuse;
MELINOE \citep{raje2026melinoe}, up to $3\times$ fewer transfers) adapts a pretrained
router for reuse rather than co-adapting experts and routing from step one.
Cache-conditional \emph{routing}
\citep{cachecond2024} is a training-free, cache-aware routing strategy that cuts mobile
cache misses $>$50\% at 0.1--3\% PPL (surpassing the oracle cache bound), and
expert-locality decode routing \citep{eldr2026} exploits locality at serving time; and
\citet{liang2026localrouting} measure this ``local routing consistency'' across 20 MoE
models, finding it trades off against load balance---corroborating why a locality
intervention carries a capacity cost. Our contribution relative
to these is the pre-registered, strict-criterion evaluation, the domain-confinement
arm, a 137M$\to$340M scale rung (the tax does not shrink), and the training-time
$\times$ training-free stacking result. \textbf{Edge MoE and expert offloading.} PowerInfer/PowerInfer-2
\citep{song2024powerinfer,xue2024powerinfer2}, EdgeMoE \citep{yi2023edgemoe}, MoE-Infinity
\citep{xue2024moeinfinity}, HOBBIT \citep{tang2024hobbit}, AdapMoE \citep{zhong2024adapmoe},
Fiddler \citep{kamahori2025fiddler}, cross-layer gate prediction (Fate;
\citealp{fang2025fate}), predictive caching with token scheduling (ExpertFlow,
predating cache-conditional routing; \citealp{he2024expertflow}), and related
systems predict, prefetch, cache, or
CPU-GPU-orchestrate experts across the hierarchy---the lineage of PMS; MoE-Infinity's
activation tracing is the direct ancestor of our tracer, and Fiddler's CPU-side expert
execution is an alternative to the disk-streaming regime we measure; see
\citet{liu2025moesurvey} for a system-stack survey of MoE inference optimisation.
\textbf{Offloaded inference and the bandwidth bound.} FlexGen \citep{sheng2023flexgen}
and ZeRO-Inference characterise memory-bound generation. \textbf{Complementary
axes we hold fixed.} Quantization \citep{frantar2023gptq,lin2024awq,ma2024bitnet158} and KV-cache
compression \citep{xiao2024streamingllm} attack bytes-per-token by changing precision or
state; speculative decoding \citep{leviathan2023speculative,li2024eagle} trades compute for
bandwidth---orthogonal to training-time routing and deliberately out of scope
here. A full, PDF-verified bibliography accompanies this draft.

\section{Conclusion}
On commodity hardware, serving a 235B MoE is a memory-bandwidth problem, and the
speed ceiling is set by bytes-per-token over bandwidth-to-bytes. We measured the
wall, built an instrument to see the routing structure inside it, and tested
whether that structure can be trained to be more cacheable. It can---but not for
free at small scale, and our pre-registered joint criteria are not met by any
configuration. We publish the negative result in full because it converts a
vague hope (``train routers to be cache-friendly'') into a precise, falsifiable
scaling claim, and because the discipline of reporting it is the point.

\paragraph{Reproducibility.} Pre-registration, training/analysis code, the
\texttt{llama-moe-trace} tool, all router traces, and per-run configs, seeds,
and data hashes are released at \url{https://github.com/Shriniwas410/cacheable-by-design}.

\paragraph{Use of AI assistance.} The author used an AI coding assistant to help
implement the experimental harness and to draft and edit the manuscript. All
experimental design decisions, the pre-registration, the runs, and the
interpretation are the author's, who takes full responsibility for the content
and has verified every reported number against the released artifacts.

\appendix
\section{Per-Domain Results}
\label{app:perdomain}
Table~\ref{tab:main} averages over domains; Table~\ref{tab:perdomain} breaks the
baseline (arm A, two seeds) and the $\lambda{=}0.05$ locality arm (two seeds)
out by domain. The perplexity tax is \emph{not} uniform: it is concentrated in
the hardest domain (prose, $+$3.0\%) and is essentially zero for the most
predictable one (code, PPL $\approx$4.7 both arms), while the cacheability gain
is large and consistent everywhere ($\sim$0.30 reuse gain in every domain). A
model with almost no perplexity headroom on prose pays for the routing
constraint there; where the task is easy (code) the router can be made sticky
for free. This domain-dependence is itself evidence for the small-capacity
explanation in \S\ref{sec:limits}.

\begin{table}[h]\centering
\caption{Per-domain metrics, baseline (A) vs.\ locality ($\lambda{=}0.05$), each
averaged over 2 seeds. ``hit@25\%'' is LRU; ``static@50\%'' is static-pin.}
\label{tab:perdomain}
\small
\begin{tabular}{llrrrrr}
\toprule
arm & domain & PPL & $\Delta$PPL & reuse & hit@25\% & static@50\% \\
\midrule
A ($\lambda{=}0$) & prose & 63.9 & --- & 0.285 & 0.432 & 0.765 \\
A ($\lambda{=}0$) & code  & 4.7  & --- & 0.384 & 0.544 & 0.722 \\
A ($\lambda{=}0$) & math  & 27.3 & --- & 0.303 & 0.486 & 0.738 \\
\midrule
B ($\lambda{=}0.05$) & prose & 65.8 & $+$3.0\% & 0.619 & 0.777 & 0.930 \\
B ($\lambda{=}0.05$) & code  & 4.7  & $+$0.0\% & 0.685 & 0.829 & 0.954 \\
B ($\lambda{=}0.05$) & math  & 27.8 & $+$1.8\% & 0.599 & 0.771 & 0.912 \\
\bottomrule
\end{tabular}
\end{table}

\section{All Runs}
\label{app:allruns}
Table~\ref{tab:allruns} lists every training run, including the 50M-token sanity
and $\lambda$-triage runs used to size the main study. The sanity/triage rows
were run before the 200M main arms and are not used in the verdict.

\begin{table}[h]\centering
\caption{Complete run log. Metrics averaged over prose/code/math.}
\label{tab:allruns}
\small
\begin{tabular}{llrrrrrr}
\toprule
run & arm & $\lambda$ & tokens & PPL & reuse & LRU hit@25\% & static@50\% \\
\midrule
a-sanity  & A & ---  & 50M  & 69.3 & 0.307 & 0.469 & 0.731 \\
b-l05     & B & 0.05 & 50M  & 69.5 & 0.644 & 0.809 & 0.957 \\
b-l01     & B & 0.01 & 50M  & 67.7 & 0.354 & 0.518 & 0.779 \\
b-l20     & B & 0.20 & 50M  & 70.1 & 0.861 & 0.855 & 0.977 \\
\midrule
a-main s1 & A & ---  & 200M & 32.0 & 0.336 & 0.494 & 0.747 \\
a-main s2 & A & ---  & 200M & 31.9 & 0.312 & 0.480 & 0.734 \\
b-l02     & B & 0.02 & 200M & 32.0 & 0.407 & 0.564 & 0.794 \\
b-l03     & B & 0.03 & 200M & 32.5 & 0.451 & 0.608 & 0.817 \\
b-main s1 & B & 0.05 & 200M & 32.6 & 0.634 & 0.788 & 0.930 \\
b-main s2 & B & 0.05 & 200M & 32.9 & 0.635 & 0.796 & 0.933 \\
c-main    & C & ---  & 200M & 32.9 & 0.470 & 0.733 & 0.991 \\
\bottomrule
\end{tabular}
\end{table}

\section{Hyperparameters}
\label{app:hparams}
\begin{table}[h]\centering
\small
\begin{tabular}{ll@{\qquad}ll}
\toprule
Model & & Optimization & \\
\midrule
$d_{\text{model}}$ & 384 & optimizer & AdamW ($\beta{=}0.9,0.95$) \\
layers & 8 & weight decay & 0.1 \\
attention heads & 6 & peak LR & $6\times10^{-4}$ \\
experts $E$ & 16 & schedule & one-cycle, 2\% warmup \\
top-$k$ & 2 & grad clip & 1.0 \\
expert $d_{\text{ff}}$ & 768 & precision & bf16 autocast \\
vocab (GPT-2 BPE) & 50257 & batch $\times$ seq & $8\times1024$ \\
total / active params & 137M / 38M & balance weight $\alpha$ & 0.01 \\
context length & 1024 & domain weight $\mu$ & 0.1 \\
\bottomrule
\end{tabular}
\caption{Hyperparameters (identical across arms; only the router-loss weights
$\lambda,\mu$ differ).}
\label{tab:hparams}
\end{table}

\section{Router Losses (exact form)}
\label{app:losses}
For a batch of routing distributions $p_{t}^{\ell}\in\Delta^{E}$ at token $t$,
layer $\ell$, with top-$k$ indicator $z_{t}^{\ell}\in\{0,1\}^{E}$:
\begin{align*}
\mathcal{L}_{\text{balance}} &= \tfrac{1}{L}\sum_{\ell} E\sum_{e} f_e^{\ell}\,P_e^{\ell},
  \quad f_e^{\ell}=\tfrac{1}{kT}\!\sum_t z_{t,e}^{\ell},\ \ P_e^{\ell}=\tfrac{1}{T}\!\sum_t p_{t,e}^{\ell} & \text{(Switch)}\\
\mathcal{L}_{\text{loc}} &= \tfrac{1}{L}\sum_{\ell}\tfrac{1}{T-1}\sum_{t\ge2}\bigl(1-\langle p_t^{\ell},p_{t-1}^{\ell}\rangle\bigr) & \text{(reuse)}\\
\mathcal{L}_{\text{dom}} &= \tfrac{1}{L}\sum_{\ell}\tfrac{1}{T}\sum_t \sum_{e\notin\mathcal{S}(d_t)} p_{t,e}^{\ell} & \text{(confinement)}
\end{align*}
Total loss: $\mathcal{L}=\mathcal{L}_{\text{LM}}+\alpha\mathcal{L}_{\text{balance}}
+\lambda\mathcal{L}_{\text{loc}}+\mu\mathcal{L}_{\text{dom}}$. The domain slice
$\mathcal{S}(d)$ for domain $d$ (of 3) is 4 exclusive experts plus 4 shared, of
$E{=}16$.

\section{Cache Simulation}
\label{app:cachesim}
We replay each router trace (per layer, an $(T,k)$ array of expert ids) through
an independent per-layer cache of capacity $c=\lceil \text{cap}\cdot E\rceil$ and
count demand misses; reported miss/token sums over layers and normalises by
$L\cdot k$ expert-loads. Policies: \textbf{LRU} (evict least-recently-used),
\textbf{LFU} (least-frequently-used over cumulative counts), \textbf{static}
(pin the top-$c$ experts by frequency in a warmup prefix, no eviction), and
\textbf{Belady} (clairvoyant: evict the resident expert whose next use is
farthest in the future---an unachievable upper bound). The same simulator scores
the frontier-model traces of \S\ref{sec:trace} and the trained-model traces of
\S\ref{sec:results}, so a single instrument spans both.

\section{Corrections Log}
\label{app:corrections}
We record every error caught during the study, per the reproducibility stance:
\begin{enumerate}\itemsep1pt
\item ``5.3\,GB/s NVMe'' was warm page-cache inflation; the cold sequential rate
is 2.4\,GB/s (used throughout).
\item Batching throughput prediction held to $B{=}8$ then broke at $B{=}32$
(paging thrash); reported as a negative result, not tuned away.
\item \texttt{ffn\_moe\_topk} is a non-contiguous argsort view; a flat copy read
garbage (every expert appearing exactly $k$ times, sub-chance reuse). Fixed with
a stride-honoring copy; a first profiling run was discarded.
\item Embedding init $N(0,1)$ made the tied-logit loss start at 373 instead of
$\sim$10.8; fixed to std 0.02.
\item A padded-batched-GEMM expert dispatch ran $2.8\times$ slower than a
per-expert loop under early-training imbalance; reverted (parity test kept),
budgets trimmed 300M$\to$200M/arm. Logged before any results were unblinded.
\item The analysis hit-rate was initially miscomputed (summed misses over all
layers but normalised by one layer's slots, giving negative rates); fixed to
$\div(L\cdot k)$.
\end{enumerate}

\section{Reproducibility Artifacts}
\label{app:repro}
Released: the frozen pre-registration (\texttt{DESIGN.md}); training, analysis,
and orchestration code; the \texttt{llama-moe-trace} tool; 7 unit tests covering
loss shapes, gradient flow, the constant-routing limits of
$\mathcal{L}_{\text{loc}}$ and $\mathcal{L}_{\text{dom}}$, the domain-mask
layout, determinism, and grouped-vs-loop dispatch parity; all router traces
(\texttt{.npz}); and per-run \texttt{config.json} with seed, loss weights, token
budget, and data-shard SHA-256 hashes. Every number in this paper is regenerable
from these. Code and artifacts: \url{https://github.com/Shriniwas410/cacheable-by-design}.

\small
\bibliography{references}
\end{document}